\documentclass[12pt,british,aps, pre, reprint, superscriptaddress, a4paper, showkeys]{revtex4-1}
\usepackage{amsmath}
\usepackage{mathptmx}
\usepackage{newtxmath}
\usepackage[T1]{fontenc}
\usepackage[latin9]{inputenc}
\usepackage[a4paper]{geometry}
\usepackage{babel}
\usepackage{verbatim}
\usepackage{units}
\usepackage{graphicx}
\PassOptionsToPackage{normalem}{ulem}
\usepackage{ulem}
\usepackage[unicode=true,pdfusetitle,
 bookmarks=true,bookmarksnumbered=false,bookmarksopen=false,
 breaklinks=false,pdfborder={0 0 0},pdfborderstyle={},backref=page,colorlinks=false]
 {hyperref}

\makeatletter

\newcommand{\lyxdot}{.}

\usepackage[warn]{textcomp}

\usepackage{microtype}
\setcitestyle{authoryear,round}
\makeatother

\begin{document}

\title{Feed-forward approximations to dynamic recurrent network architectures}

\thanks{Author's final version, accepted for publication in Neural Computation.}

\author{Dylan R. Muir}
\email{dylan.muir@unibas.ch; http://dylan-muir.com}

\selectlanguage{british}%

\affiliation{Biozentrum, University of Basel, Klingelbergstrasse 50/70, Basel 4056, Switzerland}


\date{September 15th, 2017}
\begin{abstract}
\noindent {\normalsize{}Recurrent neural network architectures can have useful computational
properties, with complex temporal dynamics and input-sensitive attractor states.
However, evaluation of recurrent dynamic architectures requires solution of systems
of differential equations, and the number of evaluations required to determine their
response to a given input can vary with the input, or can be indeterminate altogether
in the case of oscillations or instability. In feed-forward networks, by contrast,
only a single pass through the network is needed to determine the response to a given
input. Modern machine-learning systems are designed to operate efficiently on feed-forward
architectures. We hypothesised that two-layer feedforward architectures with simple,
deterministic dynamics could approximate the responses of single-layer recurrent
network architectures. By identifying the fixed-point responses of a given recurrent
network, we trained two-layer networks to directly approximate the fixed-point response
to a given input. These feed-forward networks then embodied useful computations,
including competitive interactions, information transformations and noise rejection.
Our approach was able to find useful approximations to recurrent networks, which
can then be evaluated in linear and deterministic time complexity.}{\normalsize \par}
\end{abstract}
\maketitle

\section*{Introduction}

With very few exceptions, biological networks of neurons are highly recurrent. For
an extreme example, neurons in the primary visual cortical areas in mammalian brain
make a majority of their synaptic connections between other neurons in the local
vicinity \citep{Binzegger:JournalOfNeuroscience:2004}. Recurrent networks can give
rise to complex temporal dynamics and potentially beneficial emergent computational
properties. For example, desired relationships between the activity of several neurons
can be embedded in recurrent excitatory weights \citep{Douglas1994,Hahnloser:NeuralComputation:2003,Rutishauser:NeuralCompututation:2009};
the dynamics of the network can then selectively amplify the desired representations
while rejecting noise or undesired interpretations of an input \citep{BenYishai:ProcNatlAcadSciUSA:1995,Somers:JournalOfNeuroscience:1995,Douglas:CurrentBiology:2007}.
Chaotic temporal dynamics present in reservoirs of randomly connected neurons can
be exploited to selectively detect or generate robust temporal sequences \citep{Maass:NeuralComputation:2002,Sussillo:Neuron:2009,Laje:NatureNeuroscience:2013}.

However, simulating dynamic recurrent networks to make use of their properties in
artificial systems is inconvenient for several reasons. Such simulations are non-deterministic
in terms of the time required to find an ``answer'' for a given input. This is
because the dynamics of recurrent networks, especially stochastically-generated networks,
may not be guaranteed to be stable for every input, and may indeed not be known in
advance of a simulation. Even if stable fixed-point responses exist for every finite
input, the time taken to reach these fixed points may differ depending on the input.
This issue is exacerbated by the poor fit between simulations of recurrent networks
and commodity computational architectures (i.e. CPU\,/\,GPU).

In contrast, recent successes in using feed-forward or unrolled ``recurrent'' architectures
\citep{graves_etal13,Radford_etal15} have occurred hand in hand with development
of computational systems optimised for evaluation of feed-forward networks \citep{torch2011,jia2014caffe,theano2016}.
Modern approaches for distributed evaluation of large networks \citep{tensorflow2016}
make feed-forward architectures very attractive for a range of applied computational
tasks.

Here we examine whether the known beneficial computational properties of highly recurrent
network architectures can be realised in feed-forward architectures. We take the
approach of probing recurrent networks to quantify a mapping between inputs and fixed-point
responses. We then train feed-forward networks to approximate this mapping, and compare
the information-processing abilities of the recurrent networks with their feed-forward
approximations.

\section*{Results}

\subsection*{Recurrent networks and feed-forward approximations}

\begin{figure}
\centering{}\includegraphics[width=83mm]{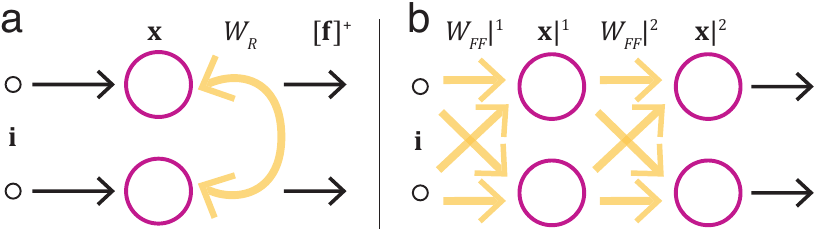}\caption{\textbf{Recurrent and feed-forward network architectures.} (a) Two-neuron recurrent
architecture. Rectified-linear (ReLU) neurons~($\mathbf{x}$; circles) receive input~($\mathbf{i}$),
and possibly reach a stable fixed point in activity ($\mathbf{f}$; the values of
network activity~$\mathbf{x}$ at the fixed point, if it exists) through recurrent
interactions via weights~$W_{R}$. (b) A $2\times2$ neuron feed-forward architecture.
Input~$\mathbf{i}$ is transformed through two layers of ReLU neurons ($\mathbf{x}|^{1},\mathbf{x}|^{2}$)
via all-to-all weight matrices~$W_{FF}|^{1}$ and~$W_{FF}|^{2}$. The activity~$\mathbf{x}|^{2}$
of layer~2 is the output of the network. See Methods for more detail. \label{fig:network-architectures}}
\end{figure}

Fig.~\ref{fig:network-architectures}a shows an example of a simple 2-neuron single-layer
recurrent network. The dynamics of each rectified-linear (or linear-threshold; or
ReLU) neuron ($x_{j}$, composed into a vector of activity $\mathbf{x}$) is governed
by a nonlinear differential equation
\begin{equation}
\tau\cdot\dot{x}+x=W_{R}\cdot\left[\mathbf{x}\right]^{+}\!+i\label{eq:linear-threshold-dynamics-results}
\end{equation}

(see Methods), and evolves in response to the input~i provided to the neuron, as
well as the activity of the rest of the network $\mathbf{x}$ transformed by the
recurrent synaptic weight matrix $W_{R}$. Here~$\left[x\right]^{+}$denotes the
linear-threshold transfer function $\left[x\right]^{+}=\max\left(x,0\right)$. Neglecting
the potentially complex temporal dynamics of network activity, for this work we define
the ``result'' of such a network as the rectified fixed-point response of the population
activity of the network~$\left[\mathbf{f}\thinspace\right]^{+}$, if a stable fixed
point exists.

In the following, we approximate the mapping between network inputs~$\mathbf{i}$
and network fixed points~$\left[\mathbf{f}\right]^{+}$ using a family of feed-forward
network architectures (Fig.~\ref{fig:network-architectures}b). For a recurrent
network with~$N=2$ neurons, the corresponding feed-forward approximation consisted
of two layers, each consisting of~$N=2$ ReLU neurons. All-to-all weight matrices~$W_{FF}|^{1}$
and~$W_{FF}|^{2}$ defined the connectivity between the network input~($\mathbf{i}$),
the neurons of layer~1~($\mathbf{x}|^{1}$), and the neurons of layer~2~($\mathbf{x}|^{2}$).
We use the notation $v|^{n}$ to refer to a variable $v$ within layer $n$. In some
implementations of unrolled recurrent network architectures, weight matrices across
several layers, representing multiple points in time, are tied together and trained
as a group. We did not take that approach with our feed-forward networks, and permitted
the weights for each layer to vary independently. The activity of $\mathbf{x}|^{2}$
were taken as the output of the network. In contrast to the recurrent network, neuron
activations in the feed-forward approximation were given by deterministic feed-forward
evaluation, with no temporal dynamics (Eq.~\ref{eq:feedforward-dynamics}; see Methods).

\subsection*{Small network architectures}

\begin{figure}
\centering{}\includegraphics[width=83mm]{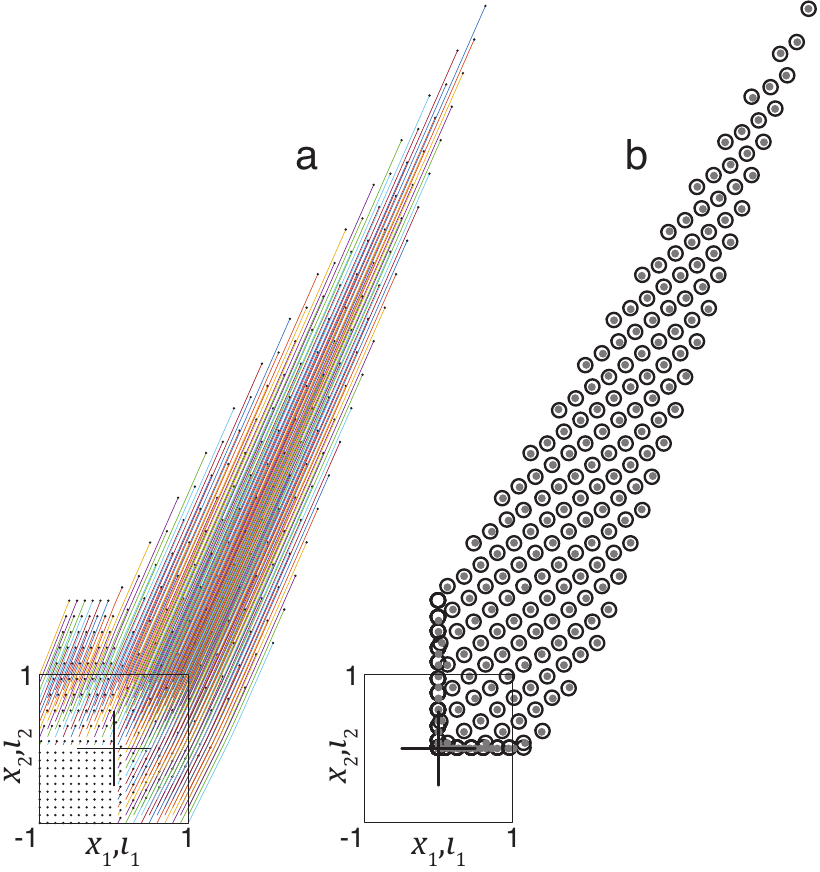}\caption{\textbf{Feed-forward approximation to the fixed-point mapping of a 2 neuron recurrent
network, with real positive eigenvalues.} (a) Recurrent dynamics for the system $W_{R}=\left[\protect\begin{array}{cc}
.4 & .2\protect\\
.8 & .5
\protect\end{array}\right]$. Loci of recurrent network responses traced to fixed points (dots), from a matrix
of inputs~$\mathcal{I}$ arranged uniformly over the unit square $\left(-1,1\right)^{2}$
(solid square). Each line traces the locus of~$\mathbf{x}$ in response to a single
input $\mathbf{i}{}_{m}$ to the corresponding fixed point $\mathbf{f}_{m}$. The
origin is indicated by the black cross. (b) Rectified fixed-point responses $\mathcal{F}\ni\left[\mathbf{f}_{m}\right]^{+}$
of the recurrent network (circles), overlaid with the corresponding feed-forward
network response (dots). Orange lines connect poorly-mapped feed-forward responses
to the corresponding recurrent fixed point. $\left\{ W_{FF}^{1},W_{FF}^{2},\mathbf{b}_{FF}^{1},\mathbf{b}_{FF}^{2}\right\} =\left\{ \left[\protect\begin{array}{cc}
4.06 & 2.20\protect\\
2.71 & 2.46
\protect\end{array}\right],\left[\protect\begin{array}{cc}
2.33 & -0.51\protect\\
0.60 & 1.20
\protect\end{array}\right],\left[\protect\begin{array}{c}
.46\protect\\
.21
\protect\end{array}\right],\left[\protect\begin{array}{c}
-2.34\protect\\
-1.77
\protect\end{array}\right]\right\} $. \label{fig:2x2-simple-dynamics}}
\end{figure}

\begin{figure}
\begin{centering}
\includegraphics[width=83mm]{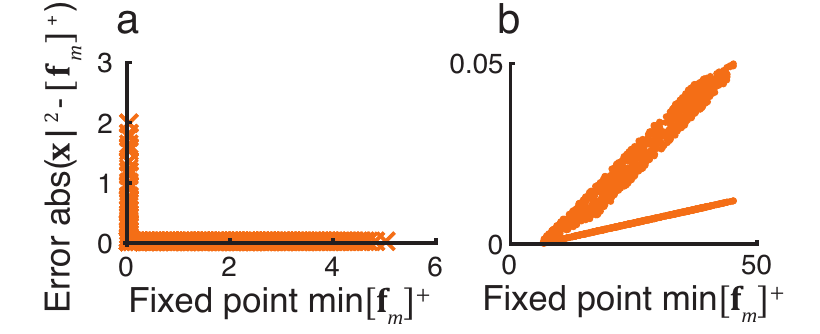}
\par\end{centering}
\caption{\textbf{Errors in the feed-forward approximation occur mainly around the activation
threshold.} (a) Large differences between the output of the feed-forward approximation
($\mathbf{x}|^{2}$) and the rectified fixed-point of the recurrent network ($\left[\mathbf{f}_{m}\right]^{+}$)
occur mostly when the activation of one recurrent unit is below threshold (i.e. $x_{1},x_{2}<0$;
$\left[\textbf{f}_{m}\right]^{+}=0$). (b) Errors in generalization increase as the
input to the network moves further outside the trained region ($\iota_{1},\iota_{2}>1$),
but remain small. \label{fig:errors-plot-2x2}}
\end{figure}

We first investigated whether the dynamics of 2-neuron recurrent networks can be
approximated by training a two-layer linear-threshold neuron (ReLU) network to directly
map network inputs to fixed-point responses of the recurrent network. We obtained
accurate feed-forward approximations for randomly chosen neuron recurrent networks
that exhibited stable non-trival fixed points. Here we show two examples of networks
with both non-oscillatory and oscillatory dynamics. Fig.~\ref{fig:2x2-simple-dynamics}
shows the result of approximating a 2-neuron recurrent network with positive real
eigenvalues, which lead to stable fixed points with an expansive mapping of the input
space.

We performed a random sampling of the input space by drawing uniform random variates
from the unit square $\left(-1,1\right)^{2}$. For each input, we analysed the eigenspectrum
and solved the dynamics of the recurrent network to determine whether a stable fixed-point
response existed for that input, discarding inputs for which no stable fixed point
existed. We therefore found a mapping between a set of inputs~$\mathcal{I}$ and
the set of corresponding fixed-point responses~$\mathcal{F}$, which was used as
training data to find an optimal feed-forward approximation to that mapping (see
Methods). Fig.~\ref{fig:2x2-simple-dynamics}a shows the activity dynamics of the
recurrent network, from a number of inputs to their corresponding fixed points.

\begin{figure}
\centering{}\includegraphics[width=83mm]{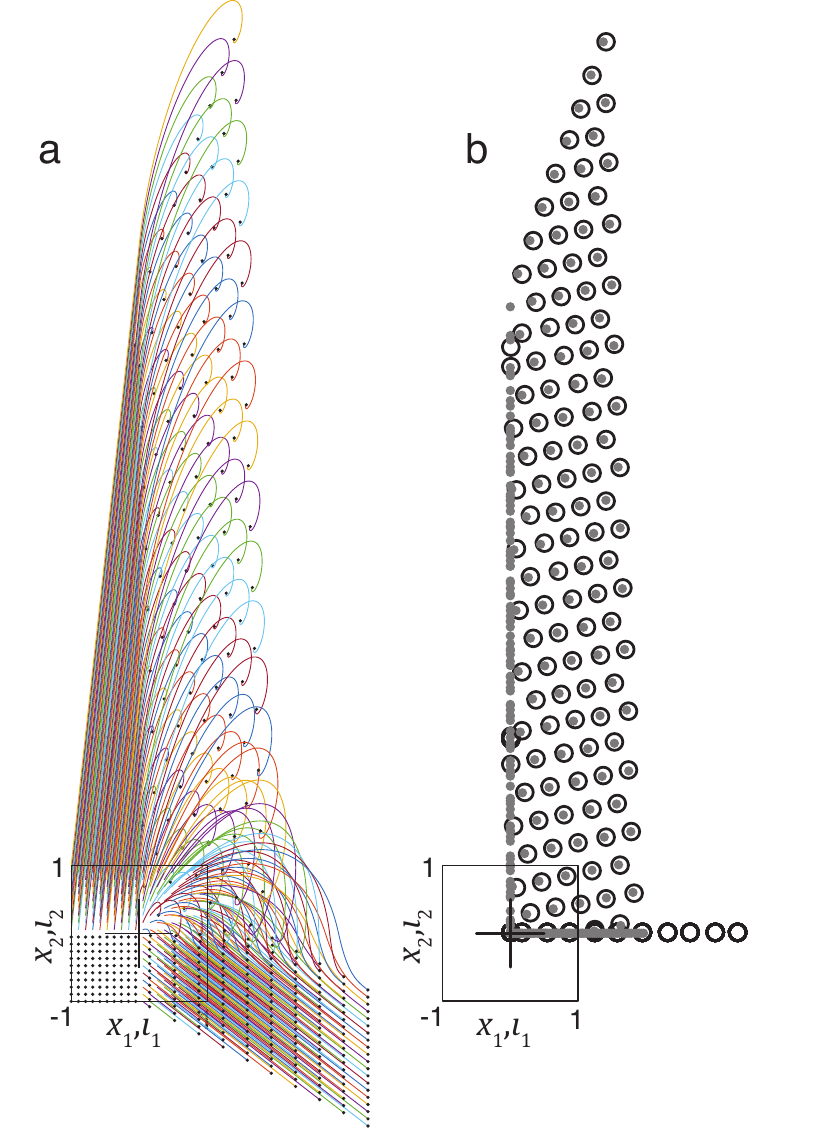}\caption{\textbf{Feed-forward approximation to a 2 neuron recurrent network with damped oscillatory
dynamics and complex eigenvalues.} (a) Recurrent dynamics for the system $W_{R}=\left[\protect\begin{array}{cc}
.70 & .11\protect\\
-.54 & .98
\protect\end{array}\right]$. This network has complex eigenvalues, with damped oscillatory dynamics. (b) Recurrent
fixed-point responses (circles) compared with feed-forward network responses (dots).
$\left\{ W_{FF}|^{1},W_{FF}|^{2},\mathbf{b}_{FF}|^{1},\mathbf{b}_{FF}|^{2}\right\} =\left\{ \left[\protect\begin{array}{cc}
1.77 & 1.29\protect\\
-0.54 & 6.89
\protect\end{array}\right],\left[\protect\begin{array}{cc}
0.23 & -0.21\protect\\
-4.35 & 1.49
\protect\end{array}\right],\left[\protect\begin{array}{c}
5.03\protect\\
0.84
\protect\end{array}\right],\left[\protect\begin{array}{c}
-10.03\protect\\
-3.19
\protect\end{array}\right]\right\} $. Notations as in Fig.~\ref{fig:2x2-simple-dynamics}. \label{fig:2x2-complex-dynamics}}
\end{figure}

We used a stochastic gradient-descent optimisation algorithm with momentum and adaptive
learning rates (Adam; \citealp{Kingma:InternationalConferenceOnLearningRepresentations:2015})
to find a feed-forward network that approximated the mapping $\mathcal{I}\rightarrow\mathcal{F}$
by minimising the mean-square loss function $c=\nicefrac{1}{2M}\Sigma_{m=1}^{M}\left(\textbf{x}_{m}|^{2}-\left[\textbf{f}_{m}\right]^{+}\right)^{2}$
(see Methods). The Adam optimisation algorithm resulted in feed-forward approximations
with smaller errors than training using direct gradient descent without momentum.
Only fixed points in which all elements $\mathbf{f}_{m}>0$ were used for training.
We found this approach to result in better approximations to recurrent fixed points.
Since many inputs map to zero fixed point responses in the recurrent network (see
Fig.~\ref{fig:2x2-simple-dynamics}a), the training process tended to over-emphasise
them, leading to a poor representation of non-zero fixed points. Training was performed
over randomly generated batches containing~$M=50$ input to fixed-point mappings,
and was halted when the batch training error smoothed over~100 batches converged.

Fig.~\ref{fig:2x2-simple-dynamics}b shows the $\mathcal{I}\rightarrow\mathcal{F}$
mapping produced by the best feed-forward network found after 16\,500~training
iterations. Inputs that lead to a non-zero response from both neurons were mapped
with high accuracy (overlapping dots and circles).

Errors in the feed-forward approximation occurred mainly around the activation threshold
(Fig.~\ref{fig:errors-plot-2x2}a). The feed-forward approximation also generalized
well for inputs outside the training regime (i.e. $\iota_{1},\iota_{2}>1$; Fig.~\ref{fig:errors-plot-2x2}b).
Generalization errors increased slowly further from the trained input space, but
remained small. In this example, we were therefore able to train an accurate feed-forward
approximation to the fixed-point dynamics of this simple recurrent network.

How does this approach fare, when applied to a recurrent network with more complex
dynamics? Fig.~\ref{fig:2x2-complex-dynamics} shows the result of approximating
a 2-neuron recurrent network with a complex eigenvalue pair with positive real part,
which leads to stable spiral fixed points. This recurrent network exhibited damped
oscillatory dynamics when driven by constant inputs (Fig.~\ref{fig:2x2-complex-dynamics}a).
Nevertheless, our approach of approximating the mapping $\mathcal{I}\rightarrow\mathcal{F}$
was successful. Fig.~\ref{fig:2x2-complex-dynamics}b shows a comparison between
the recurrent and feed-forward network mappings. As before, errors in the feed-forward
approximation were restricted to the area around the neuron activation threshold.
Our approach is therefore able to find feed-forward approximations to recurrent networks
with complex temporal dynamics.

\subsection*{Competitive networks with partitioned excitatory structure}

\begin{figure}
\centering{}\includegraphics[width=83mm]{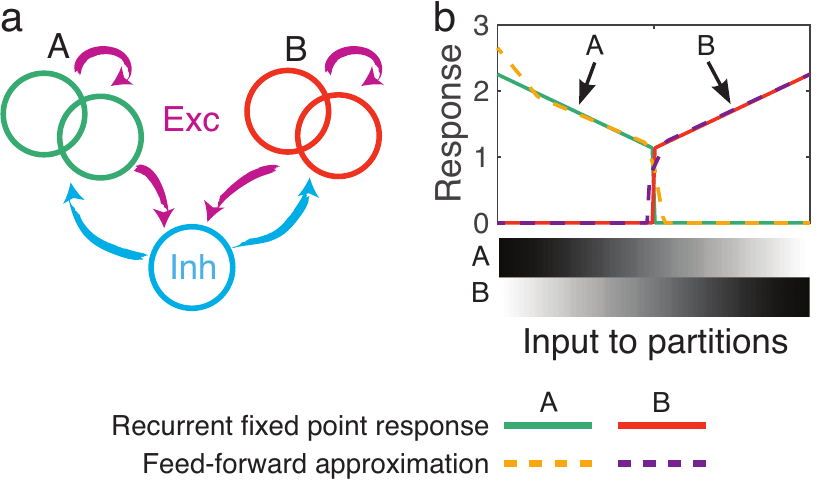}\caption{\textbf{Competition in a two-partition network.} (a)~A recurrent dynamic network
with two all-or-nothing excitatory partitions (A and B), and a single global inhibitory
neuron (Inh). The architecture of the feed-forward approximation network is as shown
in Fig.~\ref{fig:network-architectures}. (b)~Stimulating the partitions with a
linear mixture of input currents $\left(0,1\right)$ (grey shading) provokes strong
competition in the response of the partitions (note rapid switching between partition
A and B when inputs are roughly equal). The feed-forward approximation (dashed lines)
exhibits similar competitive switching to the recurrent network (solid lines). Recurrent
network parameters: $\left\{ N,w_{E},w_{I},\mathbf{b}\right\} =\left\{ 5,2.5,8,\mathbf{0}\right\} $.
\label{fig:two-partition-network}}
\end{figure}

There is growing evidence for network architectures in cortex that group excitatory
neurons into soft-partitioned subnetworks \citep{Yoshimura:Nature:2005,ko2011functional}.
Connections within these subnetworks are stronger and more prevalent \citep{Cossell:Nature:2015}.
Subnetwork membership may be defined by response similarity; neurons with correlated
responses over long periods will therefore tend to be connected \citep{ko2011functional,Cossell:Nature:2015,Lee:Nature:2016}.
These rules for connection probability and strength can give rise to network architectures
with complex dynamical and stability properties, including selective amplification
and competition between partitions \citep{Muir:PhysicalReviewE:2015}.

We investigated a simplified version of subnetwork partitioning, with all-or-nothing
recurrent excitatory connectivity (Fig.~\ref{fig:two-partition-network}a; see example
matrix in Methods). Networks with this connectivity pattern exhibit strong recurrent
recruitment of excitatory neurons within a given partition, coupled with strong competition
between partitions mediated by shared inhibitory feedback. As a consequence the recurrent
network can be viewed as solving a simple classification problem, whereby the network
signals which is the greater of the summed input to partition A ($\iota_{1+2}=\iota_{1}+\iota_{2}$)
or to partition B ($\iota_{3+4}=\iota_{3}+\iota_{4}$). In addition, the network
signals an analogue value linearly related to the difference between the inputs.
If $\iota_{1+2}>\iota_{3+4}$ then the network should respond by strong activation
of $x_{1,2}$ and complete inactivation of $x_{3,4}$ (and vice versa for $\iota_{1+2}<\iota_{3+4}$
).

We first examined the strength of competition present between excitatory partitions,
by providing mixed input to both partitions, comparing the recurrent network response
with the feed-forward approximation (Fig.~\ref{fig:two-partition-network}b). Input
was provided equally to both excitatory neurons in a partition, such that $\iota_{1}=\iota_{2}$
and $\iota_{3}=\iota_{4}$. For a given network evaluation, a single mixture was
chosen such that $\sum\mathbf{i}$ was constant. When input to partition A was weak,
input to partition B was strong, and vice versa. The recurrent network was permitted
to reach a stable fixed point for a given static input mixture, and the feed-forward
approximation was evaluated with the same input pattern.

The recurrent network exhibited strong competition between responses of the two excitatory
partitions: only a single partition was active for a given network input, even when
the input currents to the two partitions were almost equal. The feed-forward approximation
exhibited very similar competition between responses of the two partitions as the
recurrent network, also exhibiting sharp switching between the partition responses
(Fig.~\ref{fig:two-partition-network}b). In addition, the feed-forward network
learned a good approximation to the analogue response of the recurrent network, as
for the simpler networks of Figs~\ref{fig:2x2-simple-dynamics}\textendash \ref{fig:2x2-complex-dynamics}.

\begin{figure}
\centering{}\includegraphics[width=83mm]{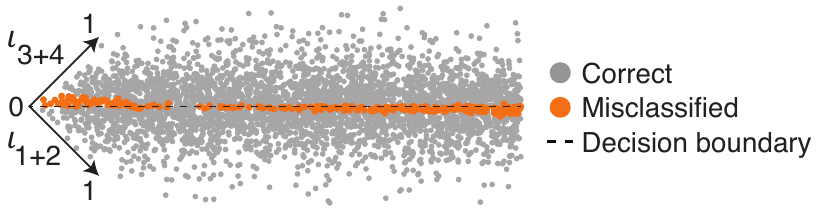}\caption{\textbf{Decision boundary is almost aligned between recurrent network and feed-forward
approximation.} Shown is the projection of the input space $\mathbf{i}$ into two
dimensions $\iota_{1+2}=\iota_{1}+\iota_{2}$ and $\iota_{3+4}=\iota_{3}+\iota_{4}$.
The ideal decision boundary of $\iota_{1+2}=\iota_{3+4}$ is indicated as a dashed
horizontal line. The majority of inputs sampled close to the decision boundary resulted
in the activation of the correct partition in both the recurrent network and feed-forward
approximation (grey dots). The decision boundary of the feed-forward approximation
was not perfectly aligned with that of the recurrent network, resulting in misclassification
of some inputs close to the decision boundary (orange dots). Network parameters as
in Fig.~\ref{fig:two-partition-network}. \label{fig:two-partition-decision-boundary}}
\end{figure}

Although the feed-forward approximation was not trained explicitly as a classifier,
we examined the extent to which the feed-forward approximation had learned the decision
boundary implemented by the recurrent network (Fig.~\ref{fig:two-partition-decision-boundary}).
Multi-layer feed-forward neural networks of course have a long history of being used
as classifiers \citep[e.g.][]{Rumelhart:Nature:1986,LeCun:NeuralComputation:1989}.
The purpose of the approach presented here is to examine how well the feed-forward
approximation has learned to mimic the boundaries between basins of attraction embedded
in the recurrent dynamic network. This question is particularly interesting for larger
and more complex recurrent networks, for which the boundaries between basins of attraction
are not known \textit{a priori}.

We examined the response of the feed-forward approximation close to the ideal decision
boundary ($\iota_{1+2}=\iota_{3+4}$; dashed line in Fig.~\ref{fig:two-partition-decision-boundary}).
We found that the majority of inputs were correctly classified by the feed-forward
approximation, but the decision boundary of the feed-forward approximation was not
perfectly aligned with the ideal, with the result that a minority of inputs close
to the boundary were misclassified by the feed-forward approximation.

\subsection*{Line attractor networks}

\begin{figure}
\centering{}\includegraphics[width=83mm]{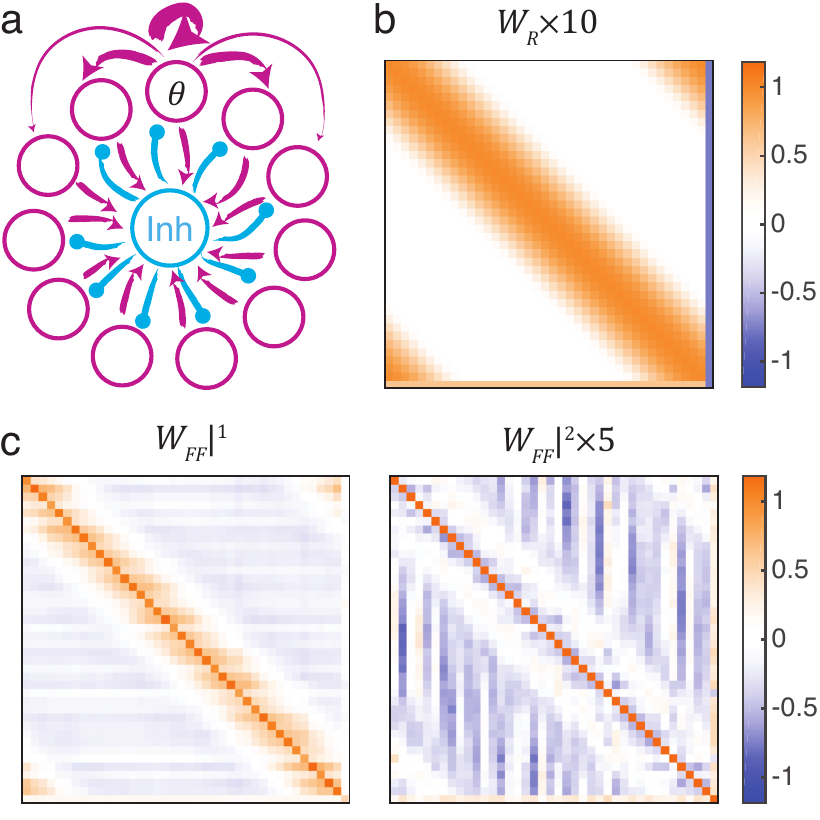}\caption{\textbf{Feed-forward approximation recovers the weight structure of a recurrent ring
model for orientation preference. }(a)~A schematic of a ring model for orientation
preference. Excitatory neurons (outer ring) are arranged in order of preferred orientation
($\theta$). Recurrent excitatory connections upper arrows are modulated by similarity
in~$\theta$ (see Methods). Inhibitory connections (from central neuron; Inh) are
made with all excitatory neurons. The architecture of the corresponding feed-forward
approximation network is as shown in Fig.~\ref{fig:network-architectures}. Recurrent
excitatory connections are shown for a single neuron, but are made identically from
all excitatory neurons. (b)~Recurrent weights $W_{R}$ implementing the ring model
in a dynamic recurrent network. Inhibitory weights were weakened by a factor of 10
for visualisation. (c)~Weights learned in a feed-forward approximation to the dynamics
of a recurrent ring model. Note the neighbourhood pattern learned by the feed-forward
network, similar to that of the recurrent network. Recurent weights $W_{R}$ and
second-layer weights $W_{FF}|^{2}$ were scaled for visualisation purposes. Recurrent
network parameters: $\left\{ N,w_{e},w_{i},\mathbf{b}\right\} =\left\{ 40,2,5,\mathbf{0}\right\} $.
\label{fig:FF-weights-ring-model}}
\end{figure}

Neurons in primary visual cortex of primates and carnivores have individual preferences
for the orientation of a line segment in visual space \citep{Hubel:JournalOfPhysiology:1962,Hubel:JournalOfPhysiology:1968};
neurons that prefer similar orientations are grouped together, and this preference
changes smoothly across the surface of cortex \citep{Bonhoeffer:Nature:1991,Blasdel:JournalOfNeuroscience:1992}.
Experimental work suggests that the sharp tuning of visual neurons for their preferred
orientation arises through recurrent processing within the cortical network \citep{Tsumoto:ExperimentalBrainResearch:1979},
rather than being defined by structured inputs to each neuron from outside the local
network \citep{Hubel:JournalOfPhysiology:1962}. The recurrent processing hypothesis
is also consistent with the fact that the majority of input synapses to each neuron
arise from other nearby neurons, and not from visual input pathways \citep{Peters:CerebralCortex:1993,Ahmed:JournalOfComparativeNeurology:1994,Binzegger:JournalOfNeuroscience:2004}.

Several recurrent network models of mammalian cortex make use of the fact that the
function of neurons changes smoothly across the surface of many cortical areas \citep{Douglas1994,BenYishai:ProcNatlAcadSciUSA:1995,Somers:JournalOfNeuroscience:1995}.
The tight relationship between physical and functional space (i.e. the preferred
orientation~$\theta$ of a neuron) suggests that local neuronal connections should
be made predominately between neurons with similar~$\theta$, falling off with distance.
In these recurrent models excitatory neurons are consequently arranged in a ring
(therefore ``\emph{ring models}''; Fig.~\ref{fig:FF-weights-ring-model}a), with
smoothly-varying~$\theta$ and with excitatory connection strength falling off with
decreasing similarity in~$\theta$. Inhibitory neurons are broadly tuned or untuned
for preferred orientation in these models, and therefore make and receive connections
with all excitatory neurons.

\begin{figure}
\centering{}\includegraphics[width=83mm]{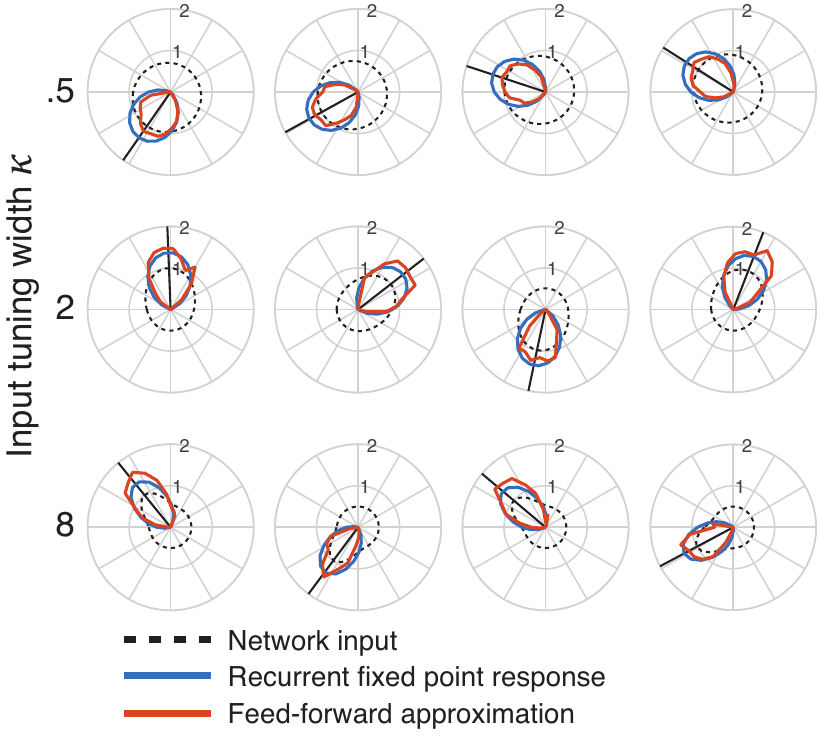}\caption{\textbf{Sharpening of broadly-tuned input.} Shown are polar plots of preferred orientation
$\theta$ versus network response amplitude, for a given input (black dashed), for
both the recurrent model (blue) and the feed-forward approximation (orange). Rows
correspond to four examples each of input under increasing tuning sharpness~$\kappa$
(indicated at left), and randomly-chosen~$\Theta$ (black line in each example).
Noise std. dev. $\zeta=0$ for these examples; common-mode input $\gamma=0.5$. All
plots have identical scaling. Note the similarity between recurrent fixed point responses
and the feed-forward approximation, and the consistency in response tuning over a
range of broadly-tuned inputs patterns. Network parameters as in Fig.~\ref{fig:FF-weights-ring-model}.
\label{fig:ring-model-sharpening}}
\end{figure}

These ring models perform powerful and useful information processing tasks, which
are supported by mechanisms of selective amplification through recurrent excitation,
coupled with competitive interactions mediated by global inhibitory feedback (also
known as \emph{winner-take all} interactions; \citealp{Douglas:CurrentBiology:2007}).
Single neurons exhibit consistent, sharp tuning for their preferred orientation~$\theta$,
in spite of poorly-tuned input. Ring networks are also able to reject significant
noise in the input, to provide a clean interpretation of a noisy signal. Recurrent
dynamics within the network establish a \emph{line attractor}, whereby a set of stable
response patterns that are translated versions of a common activity pattern are permitted
by the network.

We investigated whether a feed-forward approximation to a simple ring model for orientation
preference could capture useful information-processing features of the recurrent
network. We trained a two-layer~$40+40$ neuron network to approximate the fixed-point
recurrent dynamics of a 40-neuron recurrent ring model network (see Methods). We
generated the training mapping~$\mathcal{I}\rightarrow\mathcal{F}$ by generating
uniform random inputs~$\mathbf{i}_{m}\sim\mathcal{U}\left(.5,1\right)$ and solving
the dynamics of the recurrent network to identify the corresponding fixed points~$\mathbf{f}_{m}$
(see Methods). We discarded inputs for which no corresponding fixed point could be
found.

Fig.~\ref{fig:FF-weights-ring-model}b shows the weight matrices for the two-layer
network best approximating the recurrent dynamics, after 64\,000 training iterations.
Note that the neighbourhood relationships between similarly-tuned neurons is reflected
in the learned feed-forward weight structure, which has been acquired solely by mimicking
the fixed-point dynamics of the recurrent network (c.f. Fig.~\ref{fig:FF-weights-ring-model}c).
The locality of mapping between adjacent neuron indices was encouraged by initialising
the feed-forward weights~$W_{FF}|^{1}$ and~$W_{FF}|^{2}$ to the identity matrix
$I$ at the beginning of training (see Methods).

The ring model was designed to demonstrate how recurrent processing can lead to sharpening
of broadly-tuned inputs. To investigate whether our feed-forward approximation exhibits
similar functionality, we stimulated recurrent and feed-forward networks with broadly-tuned
inputs (Fig.~\ref{fig:ring-model-sharpening}). Indeed, the responses of the feed-forward
approximation were sharpened versions of the input, and had similar tuning sharpness
as the recurrent network. Interestingly, the sharpness of response tuning of the
feed-forward network did not change appreciably across a wide range of input tuning
sharpnesses. The feed-forward approximation was therefore able to capture the main
information-processing feature of the recurrent ring model. 

\begin{figure}
\centering{}\includegraphics[width=83mm]{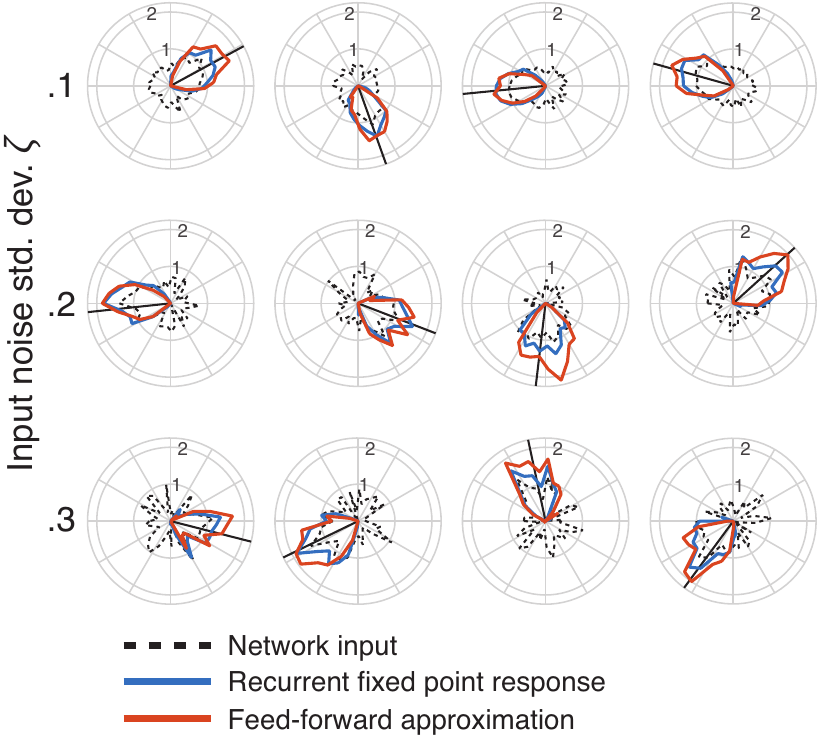}\caption{\textbf{Feed-forward approximations can mimic the noise rejection properties of the
recurrent ring model.} Rows correspond to four examples each of input under increasing
noise std. dev.~$\zeta$ (indicated at left), and randomly-chosen~$\Theta$. Tuning
sharpness~$\kappa=4$; common-mode input~$\gamma=0.5$. Network parameters and
notations as in Figs~\ref{fig:FF-weights-ring-model} and~\ref{fig:ring-model-sharpening}.
\label{fig:ring-model-noise-rejection-examples}}
\end{figure}

\begin{figure}
\centering{}\includegraphics[width=83mm]{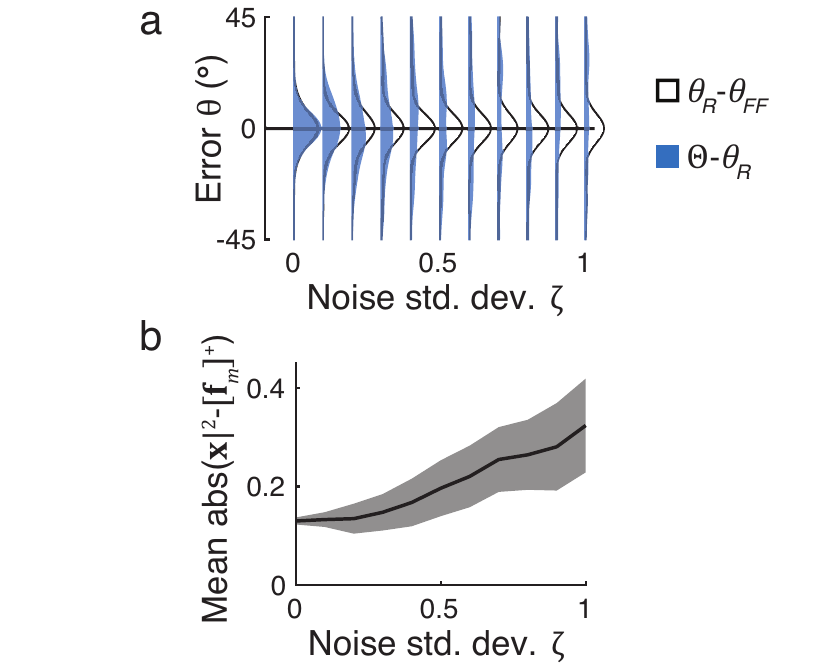}\caption{\textbf{Noise rejection by the feed-forward approximation is robust over a range
of noise amplitudes.} (a)~The error between the recurrent model angle of peak response
$\theta_{R}$ and the feed-forward approximation angle of peak response~$\theta_{FF}$
(black) was consistently small over all noise amplitudes~$\zeta$. For increasing
noise amplitudes, the ability of the recurrent model to correctly identify the orientation
of the input~$\theta_{\iota}$ degraded (blue). (b)~The accuracy of the feed-forward
approximation with respect to the recurrent model degraded gradually with increasing
noise amplitude~$\zeta$. Tuning sharpness~$\kappa=4$; common-mode input~$\gamma=0.5$.
Network parameters as in Fig~\ref{fig:FF-weights-ring-model}. \label{fig:ring-model-noise-rejection-error}}
\end{figure}

Noise rejection in the recurrent ring model is mediated by recurrent shaped excitatory
amplification of responses, coupled with global inhibitory feedback. We investigated
whether the feed-forward approximation was able to perform equivalent noise rejection,
in the absence of recurrent excitatory amplification. We stimulated the network with
tuned inputs, with increasing amounts of Normally-distributed noise with std. dev.~$\zeta$
(Figs~\ref{fig:ring-model-noise-rejection-examples} and~\ref{fig:ring-model-noise-rejection-error};
see Methods).

We quantified the error in network responses in several ways. Firstly, the purpose
of noise rejection in the recurrent model is to identify the orientation $\Theta$
of the underlying stimulus. We defined the angle of peak response $\theta_{R}$ as
the preferred orientation $\theta$ of the neuron with peak response, i.e. $\theta_{j}:j=\arg\max x_{j}$,
and defined $\theta_{FF}$ analogously for the feed-forward approximation. We then
quantified the error in stimulus interpretation between the recurrent and feed-forward
networks $\theta_{R}-\theta_{FF}$ (Fig.~\ref{fig:ring-model-noise-rejection-error}a,
black). This error was consistently clustered around zero, highlighting the closeness
of the feed-forward approximation to the behaviour of the recurrent model (see also
examples in Fig.~\ref{fig:ring-model-noise-rejection-examples}). As expected, the
ability of both models to correctly identify the underlying stimulus orientation~$\Theta$
degraded with increasing noise amplitude~$\zeta$ (increasing errors $\Theta-\theta_{R}$,
Fig.~\ref{fig:ring-model-noise-rejection-error}a, blue). The mean error between
the response of the recurrent network and the feed-forward approximation $\textrm{mean}\left(\textrm{abs}\left\{ \textbf{x}|^{2}-\left[\textbf{f}_{m}\right]^{+}\right\} \right)$
also increased with increasing noise amplitude~$\zeta$ (Figs~\ref{fig:ring-model-noise-rejection-examples}
and~\ref{fig:ring-model-noise-rejection-error}b).

\begin{figure}
\centering{}\includegraphics[width=83mm]{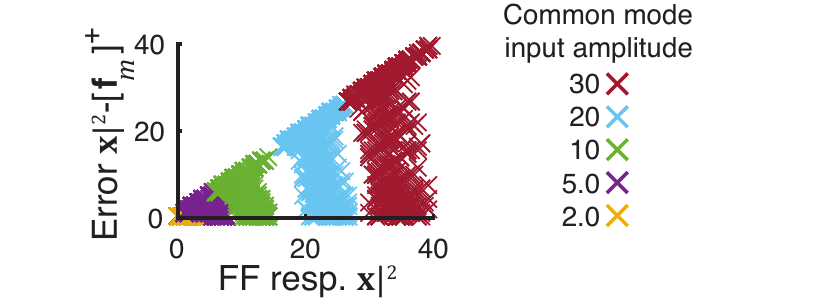}\caption{\textbf{The feed-forward approximation cannot reject common-mode input of arbitrary
amplitude.} Shown is the error between the feed-forward approximation response $\mathbf{x}|^{2}$
and the recurrent dynamic network response $\left[\mathbf{f}_{m}\right]^{+}$ to
a given input, as a function of the scale of the feed-forward response $\mathbf{x}|^{2}$,
for a range of common mode input amplitudes $\gamma$. The feed-forward approximation
was trained with $\gamma=0.5$. Errors for $\gamma<2$ were clustered close to zero.
Network parameters as in Fig.~\ref{fig:FF-weights-ring-model}. \label{fig:ring-model-common-mode-input-error}}
\end{figure}

The recurrent ring models perform common-mode input rejection, whereby the response
of the recurrent dynamic network is unchanged by adding a common-mode offset to an
input. This occurs through dynamic thresholding of the network response, provided
by global inhibitory feedback. Our feed-forward approximations were trained with
a fixed common-mode input $\gamma$ (see Methods). We examined the ability of the
feed-forward approximations to generalise their responses given arbitrarily-scaled
common mode input (Fig.~\ref{fig:ring-model-common-mode-input-error}). For feed-forward
approximations trained with $\gamma=0.5$, we found that absolute approximation errors
remained low for $\gamma\le2.0$ (i.e. error amplitudes $<1.0$). For larger $\gamma$,
errors scaled linearly with the response of the feed-forward network, indicating
that the approximation breaks down. This result suggests that matching of the input
space to the training space is required for accurate approximation, either through
appropriate selection of training inputs $\mathcal{I}$ or through input normalisation.

\begin{figure}
\centering{}\includegraphics[width=83mm]{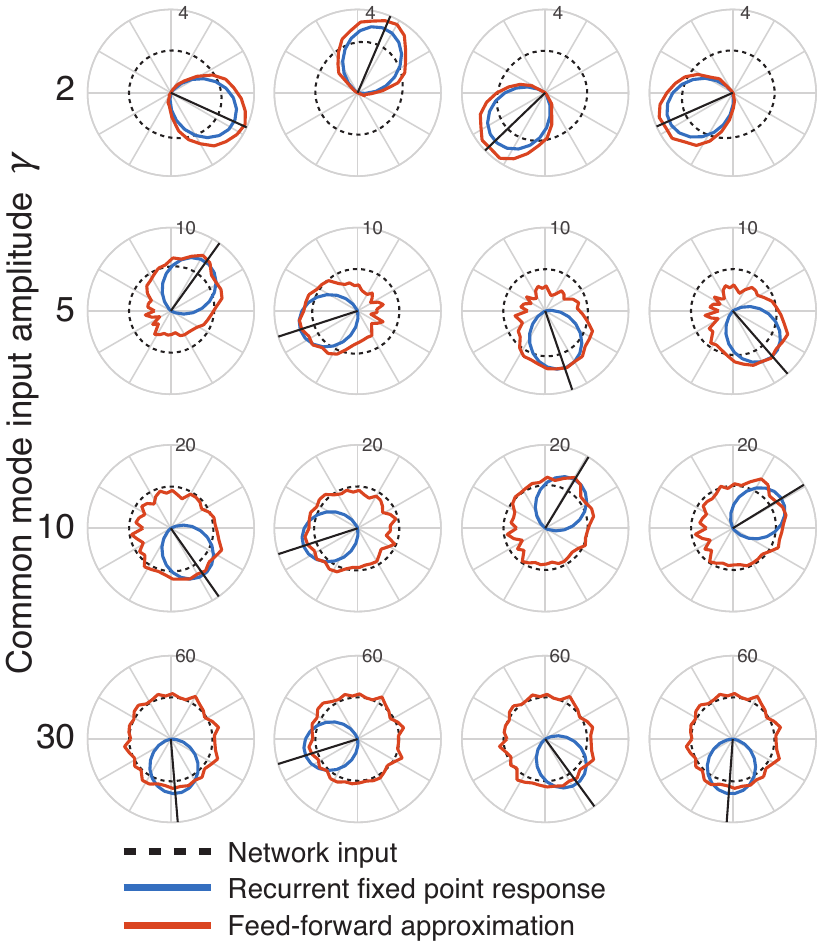}\caption{\textbf{Common-mode input rejection fails for large common-mode amplitudes.} Shown
are examples of the recurrent network (blue) and the feed-forward approximation responses
(orange) when driven with inputs with varying common-mode input amplitudes $\gamma$
(black dashed). The feed-forward approximation was trained with $\gamma=0.5$. While
the recurrent dynamic network rejected common mode inputs over all amplitudes, the
feed-forward approximation ceased to perform well for $\gamma>2$. For large $\gamma$,
common-mode noise was not rejected by the feed-forward approximation (see Fig.~\ref{fig:ring-model-common-mode-input-error}).\label{fig:common-mode-input-examples}}
\end{figure}

\section*{Discussion}

We investigated whether feed-forward neural networks could approximate the fixed-point
responses of dynamic recurrent networks. We trained two-layer feed-forward architectures
to replicate the input-to-fixed-point mapping of a dynamic recurrent networks. We
found that for small arbitrary networks, larger networks with partitioned excitatory
and inhibitory neurons, and multiple partitioned excitatory populations, as well
as even larger networks embedding line attractors, two-layer feed-forward approximations
were able to successfully reproduce the fixed-point responses of dynamic recurrent
networks. 

Feed-forward approximations reproduced the fixed-point responses for two-neuron dynamic
recurrent networks, for recurrent networks with both simple and complex temporal
dynamics (Figs~\ref{fig:2x2-simple-dynamics} and~\ref{fig:2x2-complex-dynamics}).
In the case of a dynamic recurrent network exhibiting competitive interactions between
excitatory partitions, the feed-forward approximation accurately replicated competition
between partitions (Fig.~\ref{fig:two-partition-network}). Our approach was able
to find a good approximation to a line attractor network with highly nonlinear dynamics\,\textemdash \,a
soft winner-take-all ``ring'' model for preferred orientation. This was impressive
considering that the training inputs provided to the network were uniformly randomly
distributed, and did not take into account the line attractor computation performed
by the recurrent network. The feed-forward approximation reproduced nonlinear input
transformations and noise rejection, both of which are considered to be particularly
useful features of recurrent computation in the model (Figs~\ref{fig:ring-model-sharpening}\textendash \ref{fig:ring-model-common-mode-input-error}).

We found that the accuracy of the approximations degraded close to the activation
thresholds of the feed-forward network (Fig.~\ref{fig:errors-plot-2x2}a). This
may be due to the hard loss of gradient information below the activation threshold,
in which case using units with a soft nonlinearity might alleviate this issue. However,
the feed-forward approximations to the two-neuron recurrent networks generalized
well for inputs outside the trained input space (Fig.~\ref{fig:errors-plot-2x2}b),
with errors increasing slowly but remaining low for inputs well outside the training
regime.

The feed-forward approximation to the recurrent ring model generalized well for inputs
up to a factor of 4 outside the training regime (Fig.~\ref{fig:ring-model-common-mode-input-error}).
However, the approximation broke down for inputs with larger amplitudes, in spite
of the linear transfer functions present in each neuron. Nevertheless, this restriction
simply entails the use of normalised input spaces to ensure accuracy of the approximation.

Our feed-forward approximations implicitly assume that only the fixed-point response
to an input is important, and the temporal evolution of activity to reach that fixed
point is ignored. Some modes of operation of dynamical recurrent networks explicitly
make use of chaotic dynamics to detect and generate temporal activity sequences \citep{Maass:NeuralComputation:2002,Sussillo:Neuron:2009,Laje:NatureNeuroscience:2013}.
Computations that require access to activity trajectories will of course not be possible
under the framework we proposed here. An approach might be possible where a network
was trained with step-wise approximations to the dynamics of a recurrent network,
but the purpose of our approximations was to obviate the use of iterative solutions.
Since our feed-forward networks have no temporal dynamics, they also cannot capture
complex dynamical behaviours such as damped oscillatory or limit cycle dynamics \citep{Landsman:PhysicalReviewLetters:}.

The response of the feed-forward approximations to a given input does not depend
on previous network activity, in the formulation presented here. Responses to temporal
input sequences will therefore only be accurate if the time constant of input changes
is much slower than the time constant of the dynamics of the original recurrent network,
and if complex basins of attraction are not present. Related to this point, unrolled
recurrent architectures such as LSTM networks have been employed to process discrete
temporal input sequences \citep{hochreiter1997long,liwicki2007novel}. Our feed-forward
approximations could be operated in a similar mode by augmenting the current input
$\textbf{i}\left(t\right)$ with the previous fixed-point activity $\textbf{x}|^{2}\left(t-1\right)$. 

Feed-forward approximations to dynamic recurrent systems are a powerful tool for
capturing the information processing benefits of highly recurrent networks in conceptually
and computationally simpler architectures. Information processing tasks such as selective
amplification and noise rejection performed by recurrent dynamical networks can therefore
be incorporated into feed-forward network architectures. Evaluation of the feed-forward
approximations is deterministic in time, in contrast to seeking a fixed-point response
in the dynamic recurrent network, where the time taken to reach a fixed-point response\,\textemdash \,and
indeed the existence of a stable fixed point\,\textemdash \,can depend on the input
to the network. Feed-forward approximations provide a guaranteed solution for each
network input, although in the case of oscillatory or unstable dynamics in the recurrent
network the approximation will be inaccurate. Finally, the architecture of the feed-forward
approximations is compatible with modern systems for optimised and distributed evaluation
of deep networks.

\section*{Methods}

\subsection*{Dynamic recurrent networks}

We examined dynamic networks of fully recurrently connected linear-threshold (rectified-linear;
ReLU) neurons. ReLU neurons approximate the firing-rate dynamics of cortical neurons
\citep{Ermentrout:NeuralComputation:1998}; can be mapped bidirectionally to spiking
neuron models \citep{Shriki:NeuralComputation:2003,neftci2011systematic,neftci2013synthesizing};
and have been applied successfully in large-scale machine learning problems \citep{Glorot_etal11}.

The activity of neurons in the network evolved under the dynamics 
\begin{equation}
\tau\cdot\dot{\mathbf{x}}+\mathbf{x}=W_{R}\cdot\left[\mathbf{x}-\mathbf{b}\right]^{+}\!+\mathbf{i}\mathrm{,}\label{eq:linear-threshold-dynamics-methods}
\end{equation}
where~$\mathbf{x}=\left\{ x_{1},x_{2},\dots,x_{j},\dots x_{N}\right\} ^{T}$ is
the vector of activations of each neuron~$j$; $N$~is the number of neurons in
the network; $W\in\mathbb{R}^{N\times N}$~is a weight matrix defining recurrent
connections within the network; $\mathbf{b}\in\mathbb{R}^{N\times1}$~is the vector
of neuron biases; $\mathbf{i}\in\mathbb{R}^{N\times1}$~is the vector of constant
inputs to each neuron in the network; $\tau$~is the time constant of the neurons;
and~$\left[x\right]^{+}$is the linear-threshold transfer function~$\left[x\right]^{+}=\max\left(x,0\right)$.
Without loss of generality, in this work we took~$\mathbf{b}=\mathbf{0}$ and~$\tau=1$
for the dynamic recurrent networks.

\paragraph*{Recurrent network fixed points}

Fixed points in response to a given input~$\mathbf{i}$ were defined as those non-trivial
values for~$\mathbf{x}$ such that~$\tau\cdot\dot{\mathbf{x}}=0$. We solved the
system of differential equations Eq.\ref{eq:linear-threshold-dynamics-methods} using
a Runge-Kutta (4,5) solver. With constant input provided from~$t=0$, and with $\mathbf{x}_{t=0}=\mathbf{i}$,
if no fixed-point solution was found between~$t=\left(0,161\right)$ then the corresponding
input was abandoned. We also abandoned the search if the current active partition
(i.e. the set of neurons with activity $>0$ and their associated weights) had an
eigenvalue~$\lambda^{+}$ with largest real part~$>1$, and the corresponding eigenvector~$v^{+}$
had all positive elements \citep{Hahnloser:NeuralNetworks:1998}, indicating unstable
network activity for which no stable fixed-point would be reached. For a given input~$\mathbf{i}_{m}$,
we denote the corresponding fixed point of recurrent dynamics as~$\mathbf{f}_{m}$.
Feed-forward approximations were trained to match the rectified activity of each
neuron~$\left[\mathbf{f}_{m}\right]^{+}$.

\subsection*{Recurrent network architectures}

\paragraph*{Random networks}

We generated a number of random network architectures by choosing~$W_{R}$ where
weights~$w_{ji}$ are uniformly distributed with~$w_{ji}\sim\mathcal{U}\left(-2,2\right)$,
and~$\mathbf{b}=\mathbf{0}$. We discarded any systems for which no stable fixed
points could be found. Two examples for~$N=2$ are shown in Figs~\ref{fig:2x2-simple-dynamics}\textendash \ref{fig:2x2-complex-dynamics}.

\paragraph*{Networks with modular partition structure}

We examined networks such that columns of~$W$ were either excitatory or inhibitory,
following architectures designed to be similar to mammalian cortical neuronal networks
\citep{Rajan_Abbott06,Wei12,Dwived_Sarika13}. We defined these networks to have
modular, or planted partition sub-network structure in the excitatory population
\citep{Muir:PhysicalReviewE:2015}, inspired by connectivity patterns in mammalian
cortical networks \citep{Yoshimura:Nature:2005,ko2011functional,Cossell:Nature:2015}.
An example weight matrix is given by
\begin{equation}
W_{R}=\left[\begin{array}{ccccc}
w_{E} & w_{E} &  &  & -w_{I}\\
w_{E} & w_{E} &  &  & -w_{I}\\
 &  & w_{E} & w_{E} & -w_{I}\\
 &  & w_{E} & w_{E} & -w_{I}\\
w_{E} & w_{E} & w_{E} & w_{E} & -w_{I}
\end{array}\right]\mathrm{,}\label{eq:weight-matrix-SSN}
\end{equation}
where~$\left\{ w_{E},w_{I}\right\} =\left\{ 2,4\right\} $, and unlabelled entries
of~$W_{R}$ are zero. Networks with this structure can exhibit cooperation between
neurons within a single partition, and competition between neurons in differing partitions.

\paragraph*{Networks with embedded line attractors}

In this paper we implemented a version of the classical model for orientation tuning
\citep{Douglas1994,BenYishai:ProcNatlAcadSciUSA:1995,Somers:JournalOfNeuroscience:1995},
where recurrent amplification and competition operates on weakly-tuned inputs to
produce sharply-tuned network responses. A schematic network with the architecture
described below is shown in Fig.~\ref{fig:FF-weights-ring-model}a. Excitatory neurons
were arranged around a ring, numbered~$j=\left(1,N-1\right)$. Each neuron was assigned
a preferred orientation~$\theta$ in order around the ring, with~$\theta_{j}=\left(-\pi,\pi\right)$.
Recurrent excitatory connection strength was modulated by similarity of preferred
orientation. The symmetric connections between neurons~$i$ and~$j$, for~$i,j=\left(1,N-1\right)$,
were given by~$w_{ji}=\max\left(0,\cos\left[\theta_{1}-\theta_{2}\right]\right).$
Excitatory recurrent weights were normalised such that $\Sigma_{i=1}^{N-1}w_{ji}=w_{E}$.
Excitatory to inhibitory weights are given by~$w_{Nj}=1$, $j=\left(1,N-1\right)$.
Inhibitory weights were given by~$w_{jN}=-w_{I}/N$, with~$j=\left(1,N\right)$.

Input was provided to neurons around the ring using a von~Mises-like function, given
by 
\begin{equation}
\iota_{j}=\max\left\{ 0,\exp\left[\kappa\cos\left(\theta_{j}-\Theta\right)\right]+\gamma+z_{j}\right\} \textrm{,}
\end{equation}
where~$j=\left(0,N-1\right)$; $\Theta$~is the nominal orientation represented
by a given input pattern; $\kappa$~is a distribution parameter that determines
the sharpness of the input, where~$\kappa=0$ corresponds to a uniform input and
large~$\kappa$ corresponds to a sharply-tuned input; $\gamma$ is a common-mode
input term ($\gamma=0.5$ for training); and~$z_{j}$ are Normally-distributed frozen
noise variates with std. dev.~$\zeta$, such that~$z_{j}\sim\mathcal{N}\left(0,\zeta\right)$.
Input to the inhibitory neuron~$j=N$ was zero, i.e.~$\iota_{N}=0$.

\subsection*{Feed-forward network architecture}

We trained two-layer feed-forward linear-threshold (ReLU) networks. The response
of the network was given by
\begin{align}
\mathbf{x}|^{1}= & \left[W_{FF}|^{1}\cdot\mathbf{i}-\mathbf{b}_{FF}|^{1}\right]^{+}\label{eq:feedforward-dynamics}\\
\mathbf{x}|^{2}= & \left[W_{FF}|^{2}\cdot\mathbf{x}|^{1}-\mathbf{b}_{FF}|^{2}\right]^{+}\mathrm{.}
\end{align}
The notation $v|^{n}$ indicates a variable $v$ within layer $n$ of a feedforward
network. Feed-forward networks were trained to approximate the fixed-point responses
of a given recurrent architecture. A set of random inputs~$\mathcal{I}\ni\mathbf{i}_{m}$
was generated, and a mapping found to the set of corresponding fixed-point responses~$\mathcal{F}\ni\left[\mathbf{f}_{m}\right]^{+}$,
with fixed points found as described above. Inputs for which a corresponding fixed-point
could not be found were discarded.

The network feed-forward weights~$\left\{ W_{FF}|^{1},W_{FF}|^{2}\right\} $ and
neuron biases~$\left\{ \mathbf{b}_{FF}|^{1},\mathbf{b}_{FF}|^{2}\right\} $ were
trained using the Adam optimiser\,\textemdash \,a stochastic gradient descent algorithm
incorporating adaptive learning rates and momentum on individual model parameters
\citep{Kingma:InternationalConferenceOnLearningRepresentations:2015}, with meta-parameters
set as~$\left\{ \alpha,\beta_{1},\beta_{2},\epsilon\right\} =\left\{ 10^{-3},0.9,0.999,1.5\times10^{-8}\right\} $.
The network was optimised to minimise the mean-square loss function $c=\nicefrac{1}{2M}\sum_{m=1}^{M}\left(\mathbf{x}_{m}|^{2}-\left[\mathbf{f}_{m}\right]^{+}\right)^{2}$.
Analytical parameter gradients were calculated using backpropagation of errors; zero
gradients were replaced with small Normally-distributed random values~$\mathcal{N}\left(0,10^{-5}\right)$.
Initial values for training were set to the identity matrix plus small-magnitude
uniform random variates, such that~$\left\{ W_{FF}|^{1},W_{FF}|^{2}\right\} =\mathrm{Id}\left(N\right)+\mathcal{U}\left(0,10^{-2}\right)$;
biases were initialised to~$\left\{ \mathbf{b}_{FF}|^{1},\mathbf{b}_{FF}|^{2}\right\} =0.01$.

The Matlab implementation of the Adam optimiser used in this work is available from
\uline{\href{https://github.com/DylanMuir/fmin_adam}{https://github.com/DylanMuir/fmin\_{}adam}}.

\section*{Acknowledgements}

The author thanks S\,Sadeh, M\,Cook and F\,Roth for helpful discussions.

\bibliographystyle{apalike}

\end{document}